\documentclass{article}

\usepackage{arxiv}

\usepackage{cite}
\usepackage{amsmath,amssymb,amsfonts}
\usepackage{algorithm}
\usepackage{algpseudocode}

\usepackage{graphicx}
\usepackage{textcomp}
\usepackage{xcolor}
\usepackage[percent]{overpic}
\usepackage{paralist}
\usepackage{booktabs}
\usepackage{colortbl}
\usepackage{csquotes}
\usepackage[bookmarks=false]{hyperref}
\usepackage{fancyhdr}

\usepackage{mathtools}
\usepackage{scalerel}
\usepackage{stackengine}
\usepackage{float}

\definecolor{colorA}{RGB}{ 27, 149,  25}  
\definecolor{colorB}{RGB}{252, 252, 252}  
\definecolor{colorC}{RGB}{  0, 255,   0}  
\definecolor{colorD}{RGB}{255, 195, 255}  
\definecolor{colorE}{RGB}{133, 133, 241}  

\newcommand{\optA}{\raisebox{0.05em}{\topinset{$\scaleobj{1.1}{\bullet}$}{$\scaleobj{0.8}{\bigcirc}$}{0.13em}{-0.005em}}}
\newcommand{\optB}{\topinset{$\scaleobj{0.6}{\blacktriangle}$}{$\scaleobj{1.0}{\triangle}$}{0.25em}{}}
\newcommand{\runA}{\raisebox{-0.01em}{\topinset{$\scaleobj{1.2}{\color{colorA}\bullet}$}{$\scaleobj{1.6}{\bullet}$}{0.11em}{-0.005em}}}
\newcommand{\runB}{\raisebox{0.05em}{\topinset{$\scaleobj{0.7}{\color{colorB}\blacksquare}$}{$\scaleobj{0.9}{\blacksquare}$}{0.07em}{-0.005em}}}
\newcommand{\runC}{\raisebox{0.05em}{\topinset{\rotatebox[origin=c]{45}{$\scaleobj{0.55}{\color{colorC}\blacksquare}$}}{\rotatebox[origin=c]{45}{$\scaleobj{0.75}{\blacksquare}$}}{0.125em}{-0.005em}}}
\newcommand{\runD}{\raisebox{0.05em}{\topinset{$\scaleobj{0.8}{\color{colorD}\blacktriangle}$}{$\scaleobj{1.2}{\blacktriangle}$}{0.15em}{-0.005em}}}
\newcommand{\runE}{\raisebox{0.05em}{\topinset{$\scaleobj{0.8}{\color{colorE}\blacktriangledown}$}{$\scaleobj{1.2}{\blacktriangledown}$}{0.15em}{-0.005em}}}
\newcommand{\runF}{\raisebox{-0.01em}{\topinset{$\scaleobj{1.2}{\color{cyan}\bullet}$}{$\scaleobj{1.6}{\bullet}$}{0.11em}{-0.005em}}}

\usepackage{amsthm}
\theoremstyle{definition}
\newtheorem{defn}{Definition}

\usepackage{hyperref}
\usepackage{subcaption}

\begin{document}


\title{Multiobjectivization of Local Search:\\ Single-Objective Optimization Benefits From Multi-Objective Gradient Descent}


\author{
  Vera Steinhoff\\
  Statistics and Optimization\\
  University of M{\"u}nster \\
  M{\"u}nster, Germany \\
  \texttt{v.steinhoff@uni-muenster.de} \\
  \And
  Pascal Kerschke \\
  Statistics and Optimization\\
  University of M{\"u}nster \\
  M{\"u}nster, Germany \\
  \texttt{kerschke@uni-muenster.de}
  \And
  Pelin Aspar \\
  Statistics and Optimization\\
  University of M{\"u}nster \\
  M{\"u}nster, Germany \\
  \texttt{asparp@uni-muenster.de}
  \And
  Heike Trautmann \\
  Statistics and Optimization\\
  University of M{\"u}nster \\
  M{\"u}nster, Germany \\
  \texttt{trautmann@uni-muenster.de}
   \And
  Christian Grimme \\
  Statistics and Optimization\\
  University of M{\"u}nster \\
  M{\"u}nster, Germany \\
  \texttt{christian.grimme@uni-muenster.de}
}


\maketitle

\pagestyle{plain}
\thispagestyle{fancy}
\lfoot{\vspace*{-1.25cm}\rule{\columnwidth}{0.2pt}\\\footnotesize \textcopyright2020 IEEE. 
Personal use of this material is permitted. Permission from IEEE must be obtained for all other uses, in any current or future media, including reprinting/republishing this material for advertising or promotional purposes, creating new collective works, for resale or redistribution to servers or lists, or reuse of any copyrighted component of this work in other works.\\ This version has been accepted for publication at the \textit{IEEE Symposium Series on Computational Intelligence (IEEE SSCI)} 2020.}\cfoot{}


\begin{abstract}
Multimodality is one of the biggest difficulties for optimization as local optima are often preventing algorithms from making progress. This does not only challenge local strategies that can get stuck. It also hinders meta-heuristics like evolutionary algorithms in convergence to the global optimum. 
In this paper we present a new concept of gradient descent, which is able to escape local traps. It relies on multiobjectivization of the original problem and applies the recently proposed and here slightly modified multi-objective local search mechanism MOGSA.
We use a sophisticated visualization technique for multi-objective problems to prove the working principle of our idea. As such, this work highlights the transfer of new insights from the multi-objective to the single-objective domain and provides first visual evidence that multiobjectivization can link single-objective local optima in multimodal landscapes.
\end{abstract}


\section{Introduction}

Optimization is essentially everywhere and most real-world problems are of non-linear and multimodal nature, i.e., there may exist multiple local optima that become traps for local search~\cite{preuss2015}. That is, classical local search based on gradient descent will get stuck in local optima unless restart mechanisms or search space exploration methods prevent premature convergence. Much effort has been put into this issue. Early attempts tried to make local search more flexible, e.g., by adding search points or spanning simplex structures, to discover patterns in search space and allow non-derivative descent to the optimum~\cite{nelder1965simplex}. However, local search cannot solve these problems in general. Thus, later approaches~\cite{Baudis2015} combine originally one-dimensional global search mechanisms like the STEP global search~\cite{swarzberg1994step} and a local interpolation technique proposed by Brent~\cite{brent1973} for the multivariate case. Others combine established stochastic global search mechanisms based on clustering~\cite{Kan1987} with newer elements of global optimizers~\cite{Storn1997} to gain quality improvements of solutions and to avoid finding only local optima~\cite{Pal2013}.

\newpage
In the context of global optimization, one of the most popular heuristics for finding the global optimum and for dealing with multimodal problems are population-based methods like evolutionary algorithms (EAs). Inspired by Darwinian evolution theory, this approach applies mutation, recombination, and environmental selection of good solutions in an evolutionary loop and theoretically ensures global convergence~\cite{schwefel1993evolution,beyer2001ES}. Consequently, modern global heuristics build on EAs to ensure global optimality. Two exemplary, successful, and advanced heuristics are the HCMA~\cite{Loshchilov2013} and IPOP-CMA~\cite{Yamaguchi2017}. Both extend the covariance matrix adaptation evolutionary strategy (CMA-ES) proposed by Hansen et al.~\cite{HansenCMA2003}, which improved global search capabilities of EAs significantly by adapting the mutation distribution during the algorithm's run. The first adds the STEP global line search and a surrogate model approach, while the second adapts a termination mechanism and control of the initial step-size for restart strategies within the CMA-ES.
Within all these aforementioned concepts, the global search mechanism is the driving force that steers the hybrid algorithm into the surrounding of the global optimum,
while local search is only in charge of fine-tuning the results in the basin of attraction to reach maximum precision and increase efficiency of the approach.

In contrast to the common focus on global search, we here focus  on the challenges that gradient-based local search is facing in multimodal landscapes. In that context, we propose a conceptually new approach that transfers recent ideas from the domain of multi-objective optimization towards the single-objective domain for enabling a sophisticated gradient-based local search. This approach is able to escape local traps just by (multi-objective) gradient descent and to converge to better local optima, sometimes even into the global optimum. 
Therefore, we revisit the topic of multiobjectivization~\cite{Knowles2001,segura2016using}, i.e., the reformulation of single-objective problems as multi-objective ones.
We exploit recent insights into the structure of problem landscapes of multi-objective problems~\cite{GrimmeKT2019Multimodality} and adopt the recently proposed multi-objective gradient sliding algorithm (MOGSA), a multi-objective local search strategy, to move towards the global efficient set. Interestingly, multi-objective landscape characteristics show that local optima are not necessarily traps, when we follow the multi-objective gradient and the efficient set. This provides (also in the context of single-objective optimization) the opportunity to descent from one local optimum towards another and often better local optimum. As a byproduct, we thus also provide a first visual and conceptual proof that multiobjectivization can ``link'' local optima and enable directed descent towards superior areas of the search space in single-objective optimization.

The paper is structured as follows: after providing the background of our considered problem context in Section~\ref{sec:background}, we briefly review the visualization technique for multi-objective landscapes applied here and describe our concept in Section~\ref{sec:concept}. Section~\ref{sec:evaluation} then exemplarily demonstrates our concept's working principle and algorithmic behavior compared to standard gradient local search before Section~\ref{sec:concl} concludes the paper.


\section{Background}\label{sec:background}

In the following, we aim for optimizing box-constrained continuous single-ob\-jec\-tive optimization problems, which are of the form:
\begin{equation}\label{eqn1}
\min_{l \le x \le u} f(x)
\end{equation}
with $f:\mathbb{R}^n\to\mathbb{R}$, and $x,u,l\in\mathbb{R}^n$, $u,l$ being box constraints.

As we aim to transfer, exploit, and evaluate insights from multi-objective optimization in the area of single-objective multimodal optimization, this work is part of the field of multiobjectivization research. Knowles et al.~\cite{Knowles2001} were the first demonstrating the positive effect of multiobjectivization for reducing local optima in search space, and since then, several authors followed in conducting theoretical and empirical studies on this topic. Jensen~\cite{Jensen2004} empirically showed the benefits of so-called \textit{helper-objectives}, while Neumann and Wegener~\cite{neumannWegener2008} provided theoretical results on an improved search behavior of evolutionary algorithms using one additional objective. However, other work~\cite{Handl2008,Brockhoff2007} also showed, that multiobjectivization can have positive and negative effects on search behavior. Still, the main argument for multiobjectivization is, that within a multi-objective environment, more information is available that can be exploited by algorithms for improving their search behavior. As a consequence, some authors try to use the seemingly mightier multi-objective optimizers like NSGA-II~\cite{deb2002} on these problems~\cite{Tran2013}. Others concretely report on landscapes and the existence of plateau ``networks'',
which makes evading local optima easier~\cite{GarzaFabre2015}. 

Contrary to previous research (for an extensive review, we refer to Segura et al.~\cite{Segura2013,segura2016using}), we will use a local and deterministic multi-objective optimizer to exploit the properties of multi-objective landscapes, which we identified using a recently developed visualization technique~\cite{GrimmeKT2019Multimodality}. Also in our setting the multi-objective problem (MOP) is generated by considering one additional objective function. We define this problem using a vector valued function
\begin{equation}
    F(x)=(f_1(x),  ..., f_m(x))^T\in\mathbb{R}^m
\end{equation} that shall minimize all $m$ objectives in $F(x)$ simultaneously. (Semi-)ranking of solutions is done using the dominance relation, which states that for $a,b\in\mathbb{R}^m$ $a$ dominates $b$ $(a\prec b)$, if and only if $a_i \le b_i$ for all $i \in \{1\dots m\}$ and $a_j<b_j$ for at least one $j \in \{1\dots m\}$.
Clearly, for (at least partially) contradicting objectives there is no single optimal solution value but a set of (globally) optimal trade-off solutions for which no dominating solution in search space can be found. This set is called \emph{Pareto set}. The image of this set in objective space is called \emph{Pareto front}. 

While the Pareto set and Pareto front represent global solutions, local efficient solutions are usually not in focus of research in that domain. However, there are definitions of local efficiency~\cite{grimme2019sliding}, which capture also this aspect.  

\begin{defn}
An observation $x\in \mathbb{R}^n$ is called \underline{locally efficient}, if it is not dominated by any other point in a defined neighborhood $B_x\subseteq\mathbb{R}^n$ of $x$.
\end{defn}

While the previous definition is restricted to single solutions in the multi-objective context, continuous multi-objective problems usually comprise connected local efficient sets, which need the definition of connectedness.

\begin{defn}
A set $A\subseteq\mathbb{R}^n$ is called \underline{connected} if and only if there do not exist two open and disjoint subsets $U_1, U_2\subseteq \mathbb{R}^n$ such that $A \subseteq (U_1 \cup U_2), (U_1 \cap A) \neq \emptyset$, and $(U_2 \cap A) \neq \emptyset$. Further let $B \subseteq \mathbb{R}^n$. A subset $C \subseteq B$ is a \underline{connected component} of $B$ if and only if $C \neq \emptyset$ is connected, and $\nexists D$ with $D\subseteq B$ such that $C\subset D$.
\end{defn}

With this at hand, we can finally define the local efficient sets, which are considered here.

\begin{defn}
Let $X\subseteq\mathbb{R}^n$ an open set and $x\in X$ \underline{locally efficient}. 
The set of all locally efficient points of $X$ is denoted $X_{LE}$, and each connected component of $X_{LE}$ forms a \underline{local efficient set} (of f).
\end{defn}

Local efficient points in the (here unconstrained) continuous search space fulfill the Fritz John~\cite{FritzJohn} necessary conditions: 
let $\hat{x}\in \mathbb{R}$ a local efficient point and all $m$ objective functions of $F$
continuously differentiable in $\mathbb{R}^n$. Then, there is a vector $v\in\mathbb{R}^m$ with $0 \leq v_i, i=0,...,m$, and $\sum_{i=1}^m v_i=1$, such that

\begin{equation}\label{eqn:fritzJohn}
    \sum \limits_{i=1}^m v_i \nabla f_i(\hat{x})=0.
\end{equation}
That is, in case of local efficient points the gradients cancel each other out given a suitable weighting vector $v$. 
This property is used for visualizing MO landscapes, as well as within a recently proposed MO gradient descent strategy by 
Kerschke et al.~\cite{kerschke2016towards,grimme2019sliding,GrimmeKT2019Multimodality}, which we use in the following description of our concept.


\section{Gradient Descent By Means Of Multiobjectivization}
\label{sec:concept}

\begin{figure*}[t]
    \centering
    \includegraphics[width=0.3\textwidth]{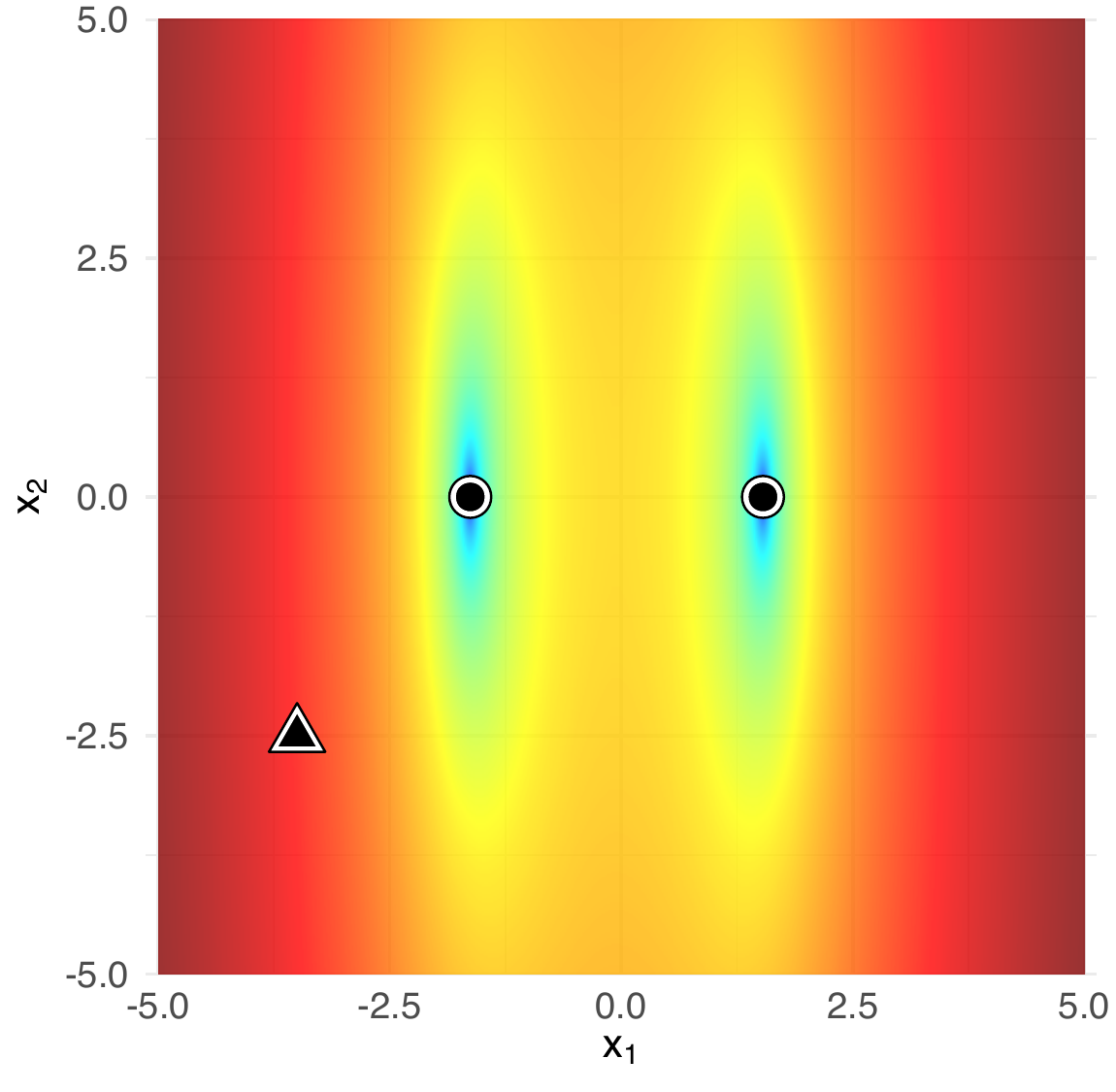}\hfill
    \includegraphics[width=0.3\textwidth]{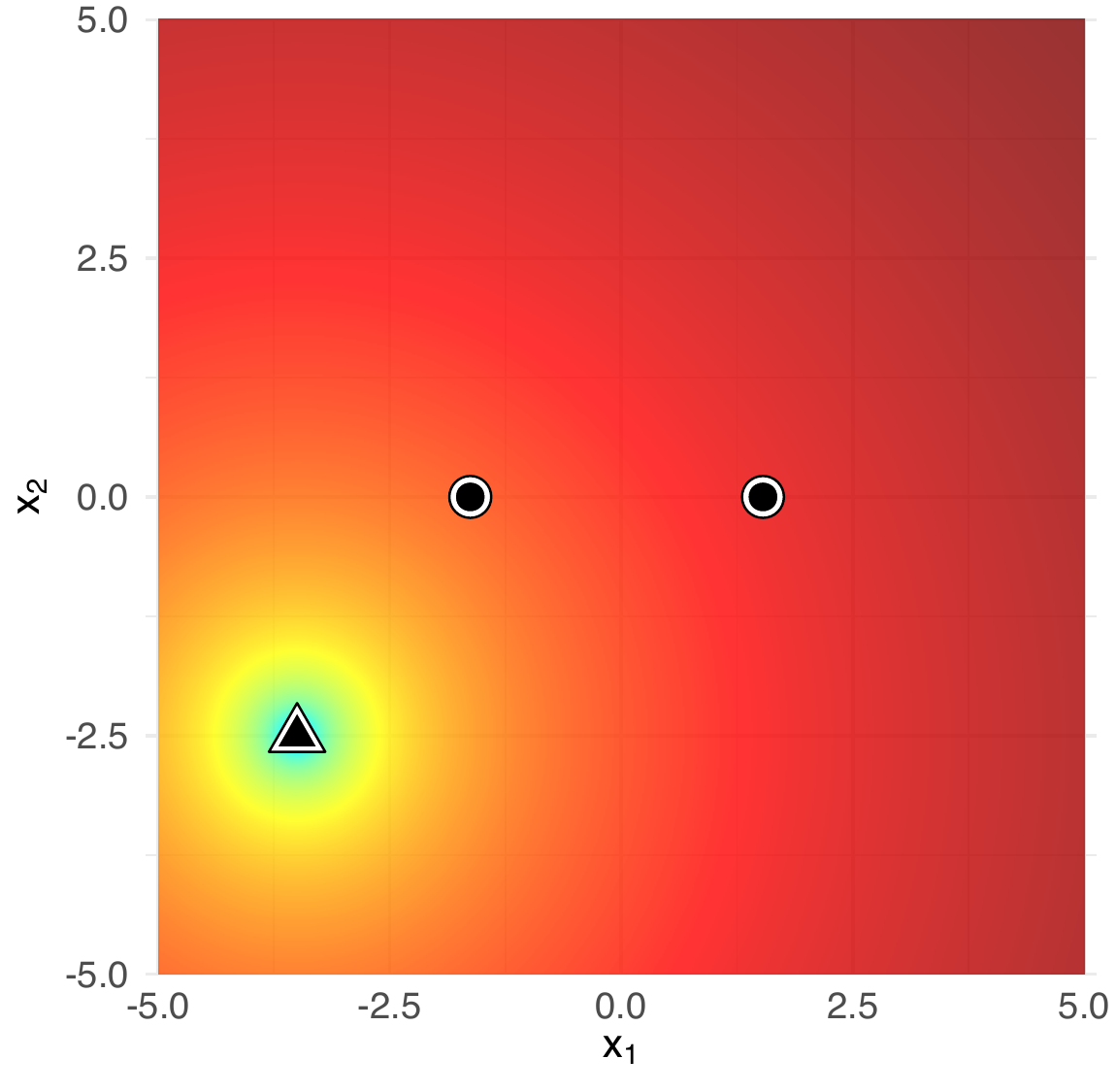}\hfill
    \includegraphics[width=0.3\textwidth]{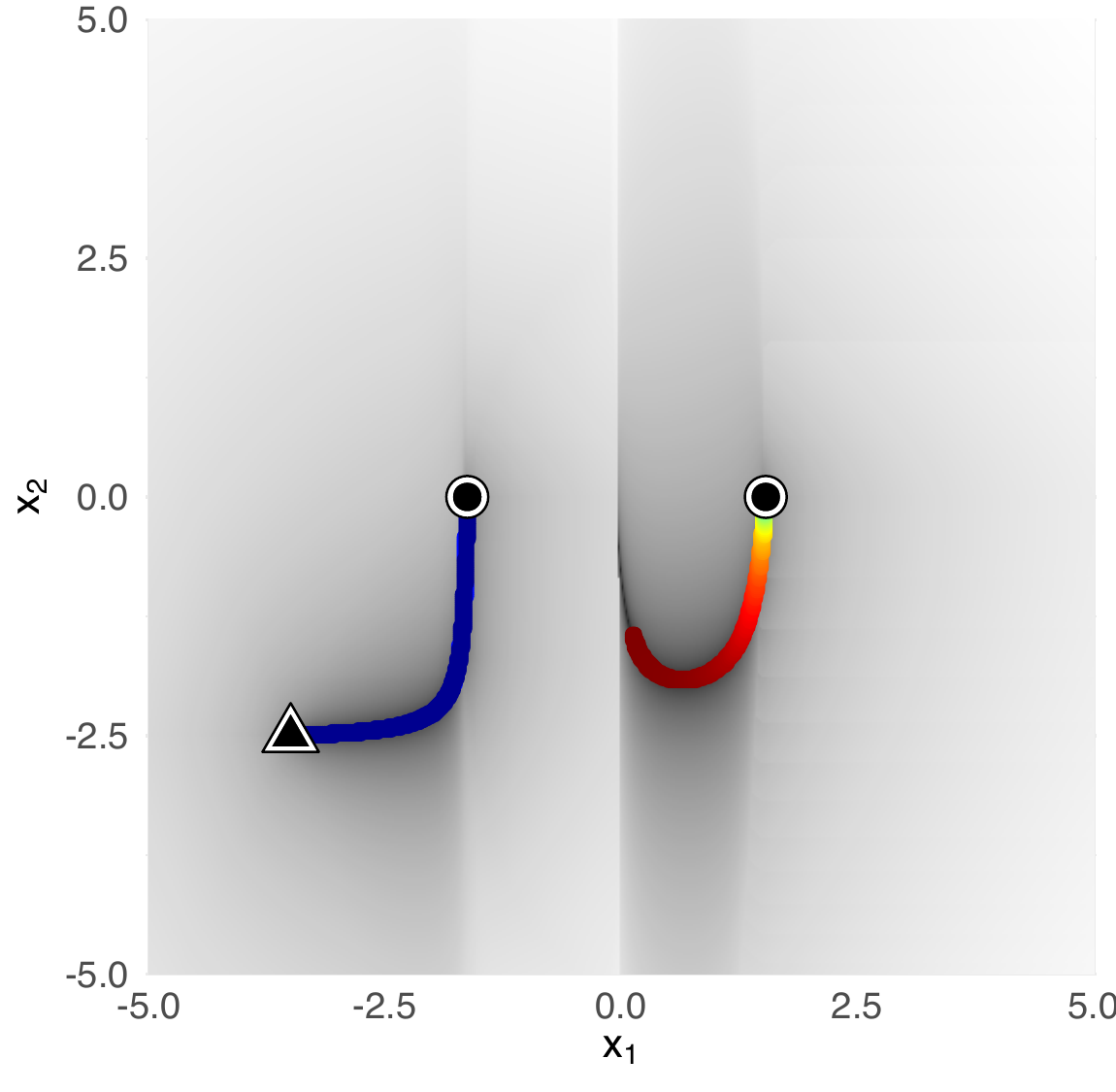}\\[0.25em]
    \includegraphics[width=0.3\textwidth]{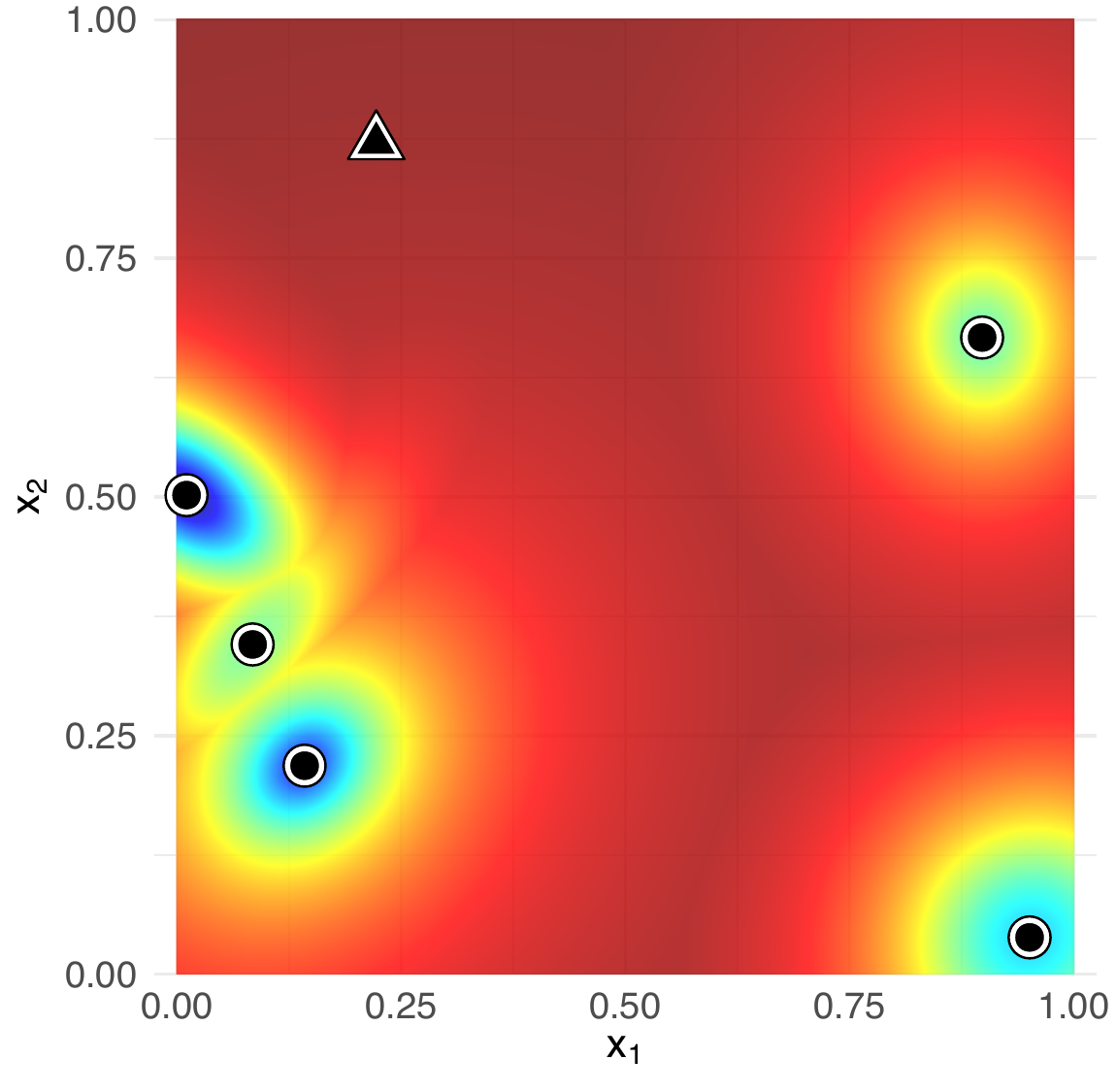}\hfill
    \includegraphics[width=0.3\textwidth]{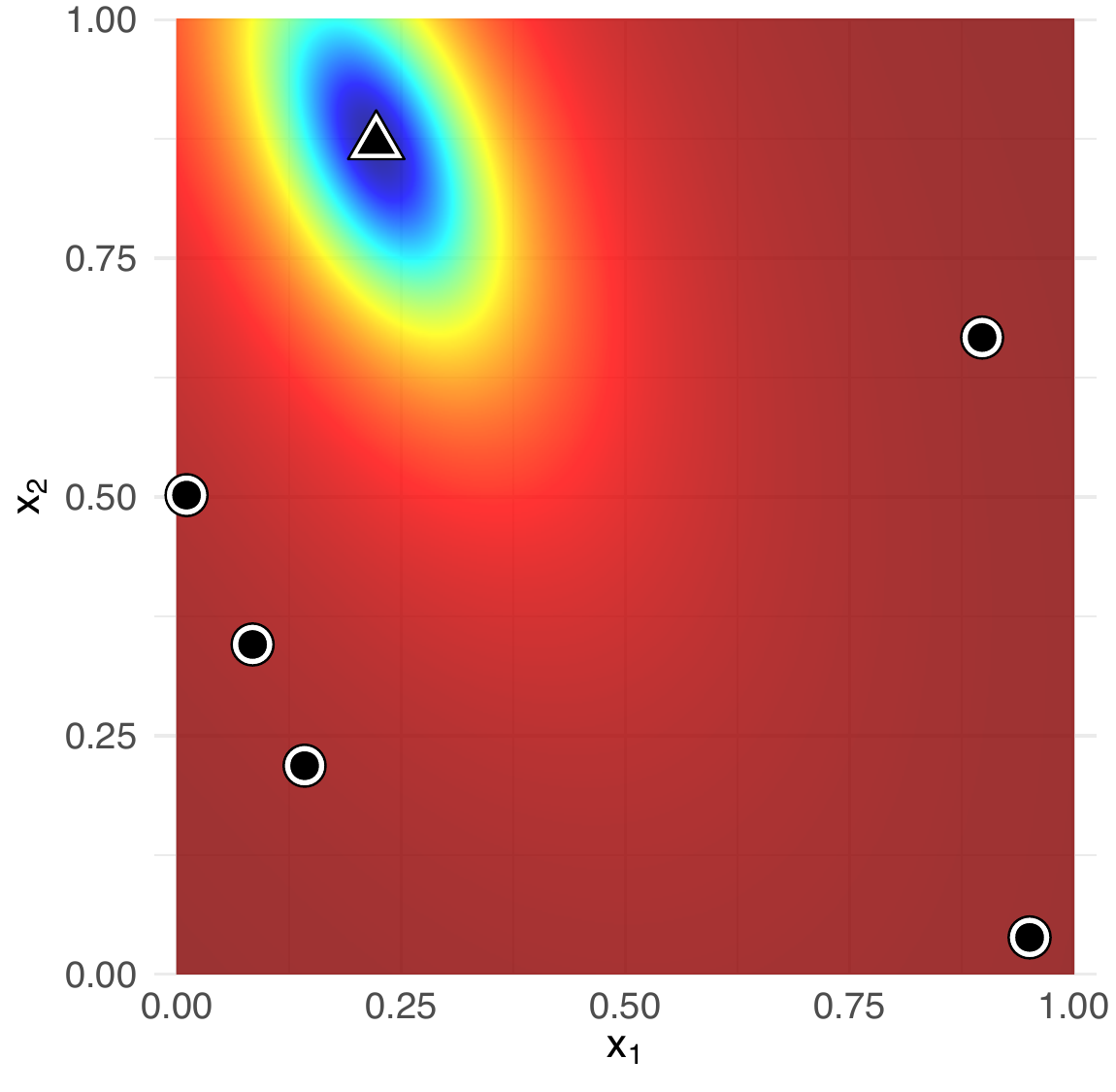}\hfill
    \includegraphics[width=0.3\textwidth]{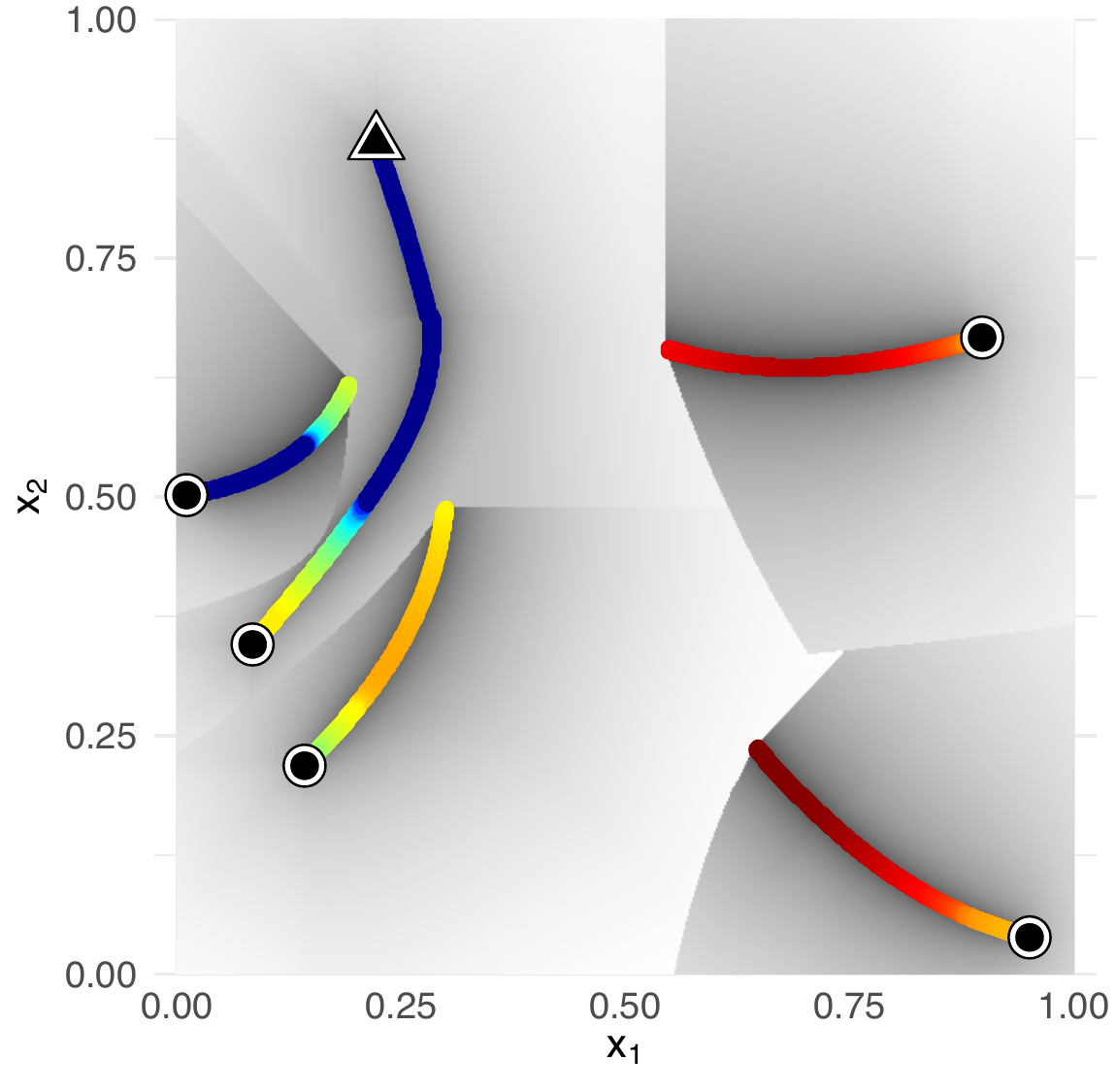}
    \caption{\label{fig:heatmaps_concept1} Exemplary combination of two single-objective problems (left column) with two spherical helper problems (middle column) to a multi-objective problem, whose landscape is visualized (right column) using the PLOT technique~\cite{SchaepermeierGK2020}. The local optima of the two single-objective problems are marked by circles (\protect\optA) and triangles (\protect\optB), respectively.}
\end{figure*}

In the following, we briefly describe the visualization of multi-objective landscapes (see Sec.~\ref{sec:visualization}). Then we detail how our approach constructs a multi-objective problem (see Sec.~\ref{sec:SOtoMO}) and how multi-objective gradient descent is adopted as local search to exploit multi-objective locality in order to reach a single-objective global optimum (see Sec.~\ref{sec:mogsa}).

\subsection{Multi-objective Landscapes Visualization in a Nutshell}
\label{sec:visualization}

For visualization of multi-objective landscapes, we use the so-called PLOT (Plot of Landscape with Optimal Trade-offs) technique recently proposed by Sch\"apermeier et al.~\cite{SchaepermeierGK2020}. This approach combines (1) a visualization of (locally) efficient solutions and their attraction basins w.r.t. multi-objective gradient information~\cite{GrimmeKT2019Multimodality} as well as (2) global quality information based on the dominance count of each solution~\cite{Fonseca1995}.

In order to visualize the multi-objective landscape for two objectives, the Fritz John condition (see Eqn.~\ref{eqn:fritzJohn}) is used. After the decision space has been discretized into regular cells, for each cell center point $\mathbf{x}$, the sum of the normalized gradients 
\begin{align}
    \nabla f(\mathbf{x}) = \nabla f_1(\mathbf{x}) / ||\nabla f_1(\mathbf{x})|| + \nabla f_2(\mathbf{x}) / ||\nabla f_2(\mathbf{x})||
    \label{eq:mog}
\end{align}
represents the multi-objective gradient descent direction for the respective cell. 

Obviously, the multi-objective gradient either points to a neighboring cell or has an approximate length of zero. The latter happens in (or near) any locally efficient point. For these locally efficient cells, we assume a ``height'' value of zero and compute the ``height'' of all other cells as accumulated length of the multi-objective gradients that describe the multi-objective gradient descent path from the respective cell to the attracting local efficient cell\footnote{Analogous to single-objective gradient descent, we may imagine this multi-objective descent path as the path, which a "multi-objective ball" would roll down towards the attracting efficient set.}. Applying a gray-scale coloring to all cells w.r.t. their ``height'' value (light gray far from and dark gray near to local efficient point) provides an intuitive notion of attraction basins of local efficient sets.

In a second step, the visualization of the locally efficient sets and the basins of attraction is augmented with additional information on the relation of efficient points w.r.t. dominance (as defined in Sec.~\ref{sec:background}). For each cell, which is considered locally efficient, PLOT determines the 
domination count regarding all other locally efficient cells. This relation is visualized with another color schema: dark blue for non-dominated (i.e., global) efficient solutions, dark red for most dominated local efficient solutions. 
An exemplary PLOT is shown in Figure~\ref{fig:heatmaps_concept1}, right column. Therein, the gray basins of attraction as well as the colored efficient sets are shown for two simple bi-objective problems.

\subsection{Multiobjectivization Procedure and Benefits of Landscape Characteristics}\label{sec:SOtoMO}
A precondition for using multi-objective techniques in the single-objective domain is to transform any considered single-objective function $f_1$ into a multi-objective problem. Therefore, we introduce a second objective $f_2$. For maximum reduction of complexity and to ensure accessible visualizations of the MO landscapes~\cite{MOlandscapes}, we add an unimodal $n$-dimensional sphere function $f_2(x)=\sum_{i=1}^n (x_i-s_i)^2$ with optimum $s \in \mathbb{R}^n$. 
Note that for the optimization process, neither a costly evaluation of the second objective $f_2$, nor a (probably also costly) approximation of its known derivative $\nabla f_2(x)=\sum_{i=1}^n 2\cdot (x_i - s_i)$, is necessary. The additional objective $f_2$ only serves as helper objective to create a MOP and with it a multi-objective landscape with all its characteristics that can be visualized and algorithmically exploited by replacing single-objective local optima with efficient sets that should guide the local search to a better region.

In Figure~\ref{fig:heatmaps_concept1} we provide a visual description of the  mul\-ti\-ob\-jec\-ti\-vi\-za\-tion procedure and the properties of the resulting MOPs using a bimodal problem and a multimodal problem respectively (left column), which are combined with spherical helper objectives (middle column). The resulting efficient sets are visualized in the right column PLOT graphic.

\begin{figure}[t]
\begin{minipage}[t]{0.45\textwidth}
 	\centering
 	\includegraphics[width=\textwidth]{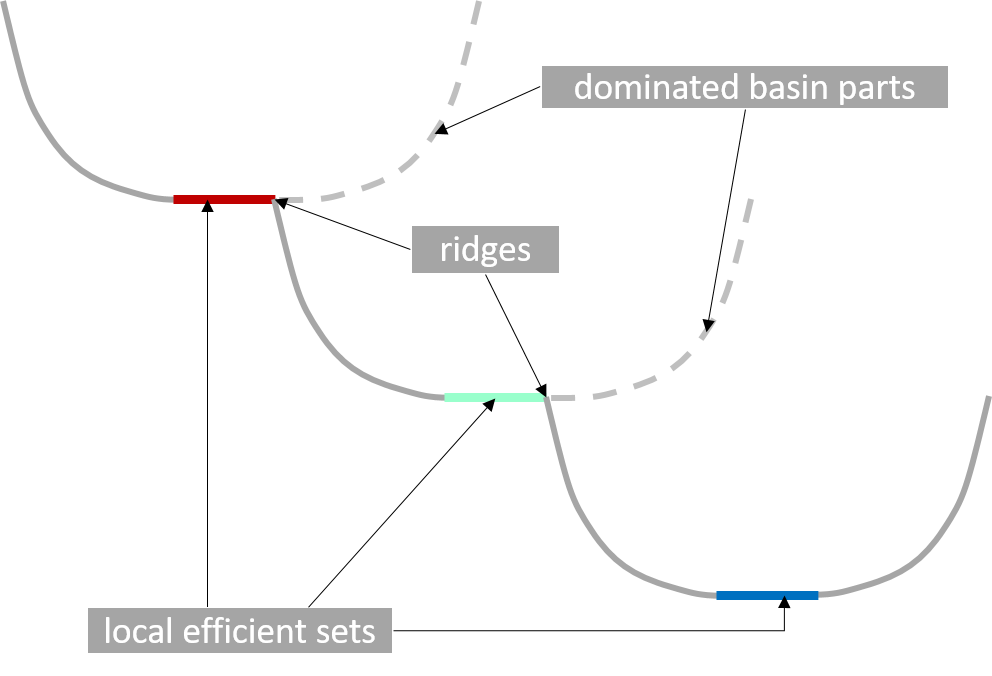}
 	\caption{\label{fig:vis_schema}Schematic depiction of superposition of attraction basins. Local efficient sets can be cut by dominating basins. Domination of basins leads to ridges in visualization. These ridges denote rapid change of attraction. Note that domination is shown as height for better interpretability only.}
\end{minipage}
\hfill
\begin{minipage}[t]{0.53\textwidth}
	\centering
	\includegraphics[width=\textwidth]{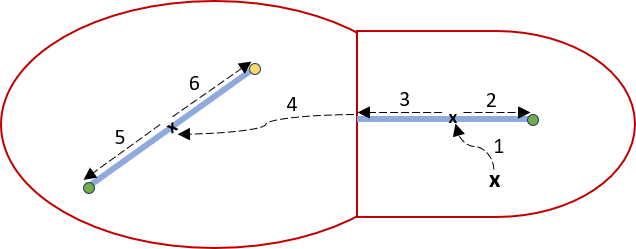}
	\caption{Schematic view on the search space of a bi-objective problem with two attraction basins (encircled in red). The optima of both single-objective functions are indicated by (green and yellow) dots, located on two efficient sets (blue lines). The dashed arrows display the search behavior of MOGSA starting in point x.}
	\label{fig:wayMOGSA}
\end{minipage}
\end{figure}

Most interesting, the multi-objective global and local efficient sets, as well as their surrounding attraction basins reveal the interaction of the objectives. 
If we observe the domination relation of the efficient sets (we can interpret this from the colors) and their respective basins of attraction, we find, that the basins superpose each other. The borders of the superposition are visible as ridges that seem to cut the local efficient sets abruptly.
In a schematic (and one-dimensional) depiction, Figure~\ref{fig:vis_schema} illustrates this superposition. 

We exploit this specific property of ridges which cut efficient sets. The respective local efficient sets can be interpreted as direct path to superposing basins which in turn are candidates for containing better single-objective local (or even global) optima. As such, the locally efficient sets can be utilized as connecting ``slides'' from one local optimum of $f_1$ to another one.

\subsection{Multi-objective Gradient Search and Single-objective Exploitation}\label{sec:mogsa}
In order to directly follow the information of the multi-objective gradient (towards the local efficient set) and to slide along the local efficient set until a potential ridge has been crossed, we apply the recently proposed multi-objective gradient sliding algorithm (MOGSA) for local optimization in the multi-objective domain~\cite{grimme2019sliding}.

MOGSA is capable of exploiting the properties of MO landscapes in two repeating phases -- see Figure~\ref{fig:wayMOGSA} for an exemplary bimodal function $f_1$ (whose optima are depicted in green) and an unimodal function $f_2$ (yellow optimum). The blue lines illustrate the efficient sets, which are located in different basins of attraction (indicated by red borders).
The local efficient set and its associated basin (on the right) are cut by a ridge. Moving across the ridge would direct a multi-objective gradient-based search towards the global efficient set. In our example, the global efficient set and also the global optima of both single-objective functions are located in the left basin of attraction.

Starting in point $x$, MOGSA follows the MO gradient to find a point on the local efficient set (first phase) in the respective basin (1). From there, it follows the single-objective gradient (second phase) of the first function $f_1$ until the (green) optimum is reached (2). The latter phase is repeated for objective $f_2$ until the end of the set is reached and a ridge has been passed (3). With this, the second phase stops and the first one is started again, searching for the efficient set in the new basin of attraction (4).

\begin{algorithm}[t]
  \small
  \caption{SO-MOGSA (naive implementation, stores complete search path up to $f_2$)}
  \label{alg:somogsa}
  \begin{algorithmic}[1]
    \Require{\textbf{a)} start point $x_s\in \mathbb{R}^n$, \textbf{b)} function $f_1$ to be optimized \textbf{c)} termination angle $t_\angle\in [0,180]$ for switching to local search w.r.t.~$f_1$, \textbf{d)} step size $\sigma_{MO}\in \mathbb{R}$ for MO gradient descent, \textbf{e)} step size $\sigma_{SO} \in \mathbb{R}$ for SO gradient descent w.r.t.~$f_2$, \textbf{f)} $y^*\in\mathbb{R}^n$ optimum of $f_2$}
    
    \State $f_2(x) = (x_1-y^*_1)^2 + \dots + (x_n-y^*_n)^2$ \Comment{use parameterized fixed sphere function for multiobjectivization}
    \State $x = x_s$
    \State $p = []$ \Comment{initialize archive for storing search path}
    \While{optimum of $f_2$ not yet     reached}
        
        \While{$|\nabla f_1(x)|> 0$ and  $\angle(\nabla f_1(x),\nabla f_2(x)) \le t_\angle$} 
            \State $x = x - \sigma_{MO}\cdot \left(\frac{\nabla f_1(x)}{|\nabla f_1(x)|} + \frac{\nabla f_2(x)}{|\nabla f_2(x)|} \right)$ \Comment{MO gradient descent}
            \State $p.store(x)$
        \EndWhile
            \State $x^{t-1} = x = LocalSearch(x, f_1)$ \Comment{local search (here gradient descent) w.r.t. $f_1$}
            \State $p.store(x)$
                \While{$\angle(\nabla f_1(x),\nabla f_2(x)) \ge 90^{\circ}$ and $\angle(\nabla f_2(x^{t-1}),\nabla f_2(x)) \le 90^{\circ}$} 
                    \State $x^{t-1}= x$
                    \State $x = x - \sigma_{SO}\cdot \frac{\nabla f_2(x)}{|\nabla f_2(x)|}$ \Comment{gradient descent towards $f_2$}
                    \State $p.store(x)$
                \EndWhile

    \EndWhile 
    \State \Return{$p$}    
    \end{algorithmic}
\end{algorithm}

With few extensions this searching principle of MOGSA can be adopted to extract the necessary information for single-objective optimization, resulting in SO-MOGSA as shown in Algorithm~\ref{alg:somogsa}. In the multiobjectivized setting, the property of MOGSA ``sliding'' to another efficient set is helpful, as a local optimum of $f_1$ is automatically also part of a local efficient set of $F$, and the global optimum of $f_1$ is part of a global set. Thus, we extend MOGSA to address two aspects: finding a precise-as-possible approximation of each local set's endpoints for $f_1$ (see line 8) using additional single-objective local search, and storing all investigated solutions of $f_1$ for later selection of the best solution (ref. to lines 3, 7, and 13). This is repeated, until MOGSA has terminated and (hopefully) reached the global set (that is the optimum of $f_2$, see line 4) -- and thereby possibly also visited the global optimum of $f_1$. Note again, as we construct the multi-objective problem from the original (black-box) single-objective problem $f_1$ and the predefined sphere function $f_2$, the execution of SO-MOGSA is computationally cheaper than in the multi-objective case. Because the gradient of $f_2$ does not have to be approximated, these costs only occur for approximating the gradient information of $f_1$.


\section{Evaluation of the Concept}
\label{sec:evaluation}

In order to demonstrate and evaluate the working principle of SO-MOGSA, we present first experimental insights into the algorithm's behaviour on well-known multimodal test problems. For multiobjectivization of $f_1$ as used in SO-MOGSA, we add a sphere function as second objective $f_2$ (see Section~\ref{sec:SOtoMO}) and fix its optimum at $(-3.5,-2.5)$. As this work is intended as validation of a new idea rather than an extensive performance study, we concentrate on visualizing the algorithm behaviour compared to a classical Nelder-Mead local search~\cite{nelder1965simplex} starting at six different points distributed in search space.

\begin{figure*}[t]
    \centering
    \includegraphics[width=0.325\textwidth]{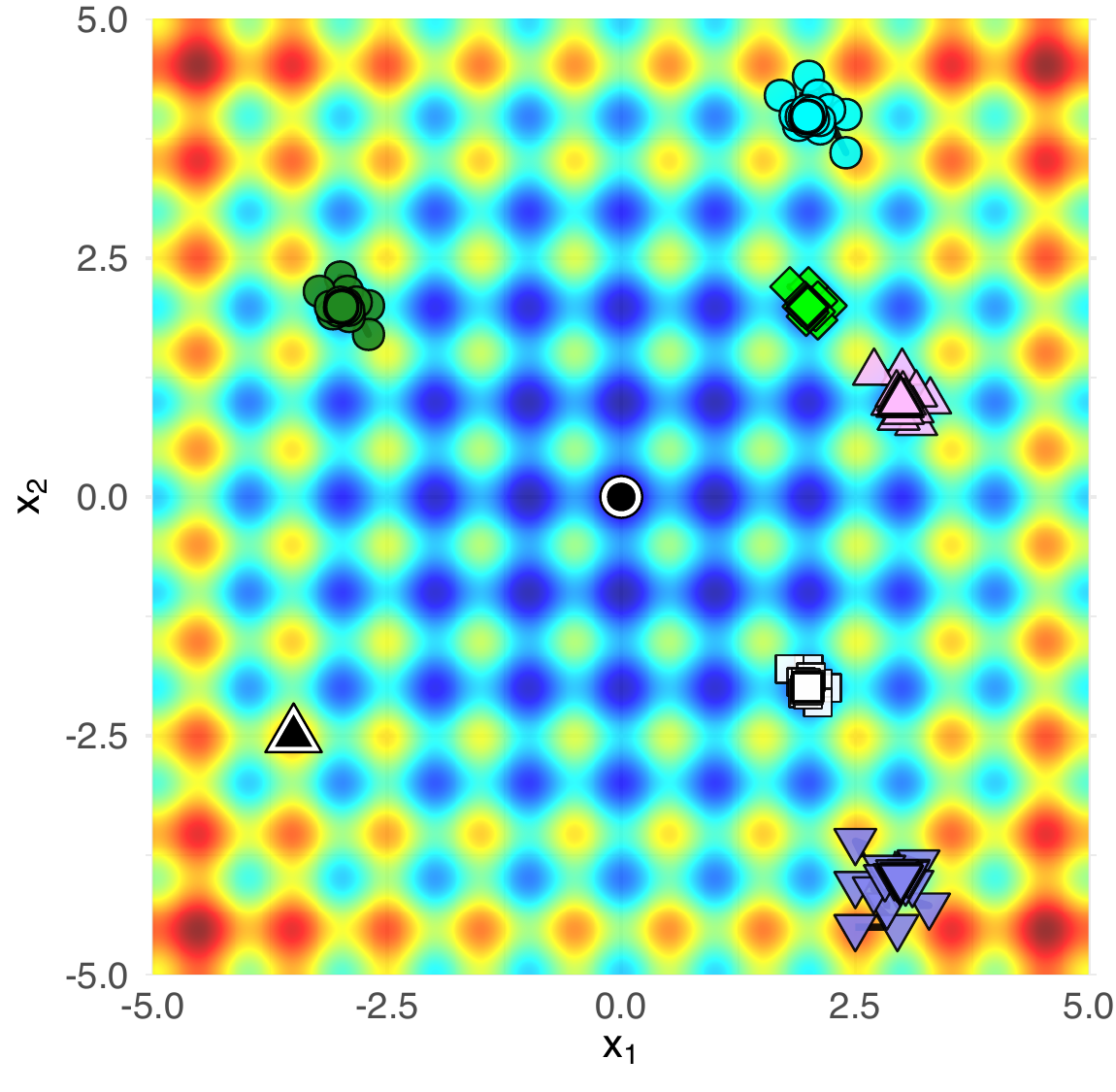}\hfill
    \includegraphics[width=0.325\textwidth]{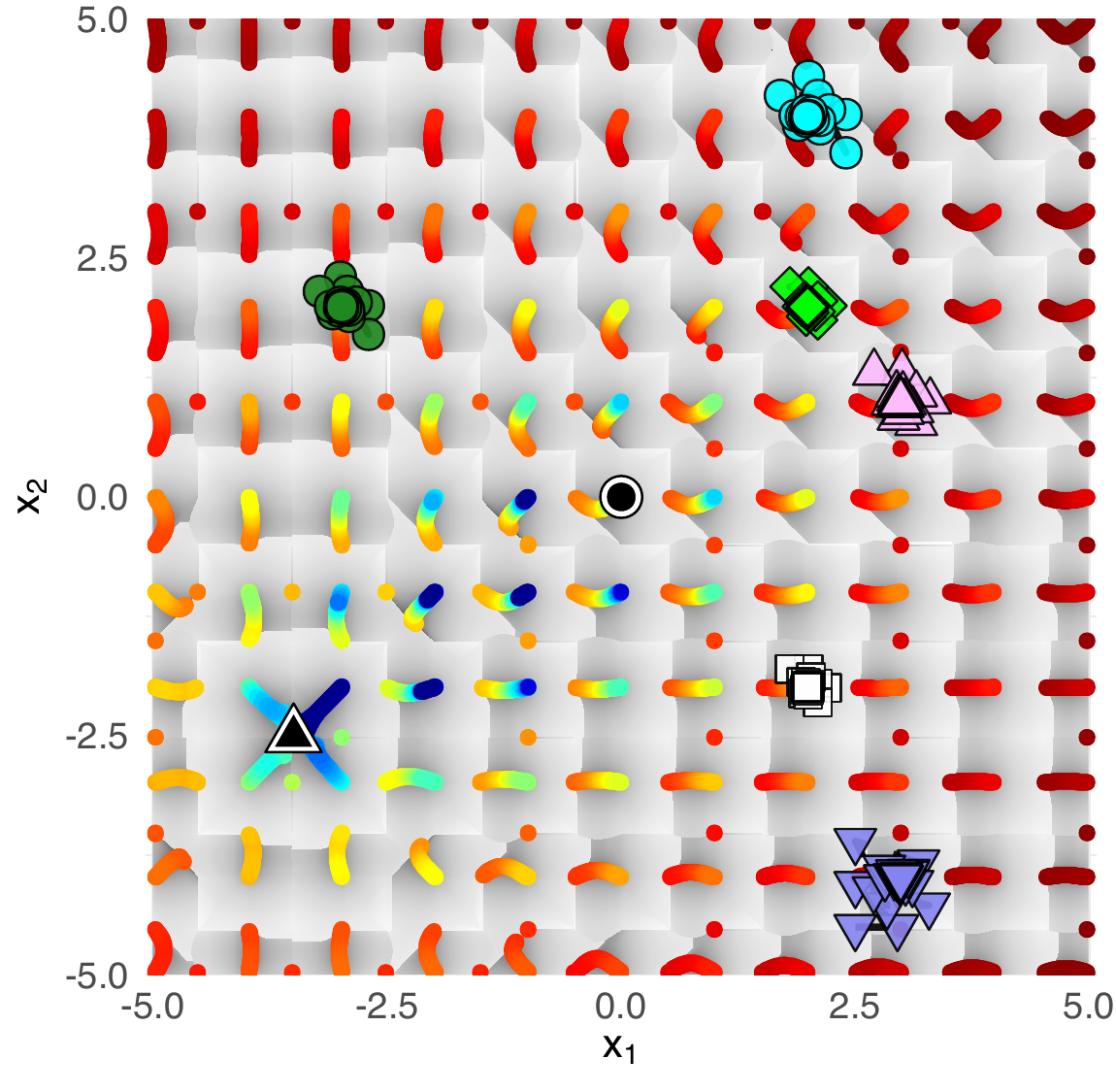}\hfill
    \includegraphics[width=0.325\textwidth]{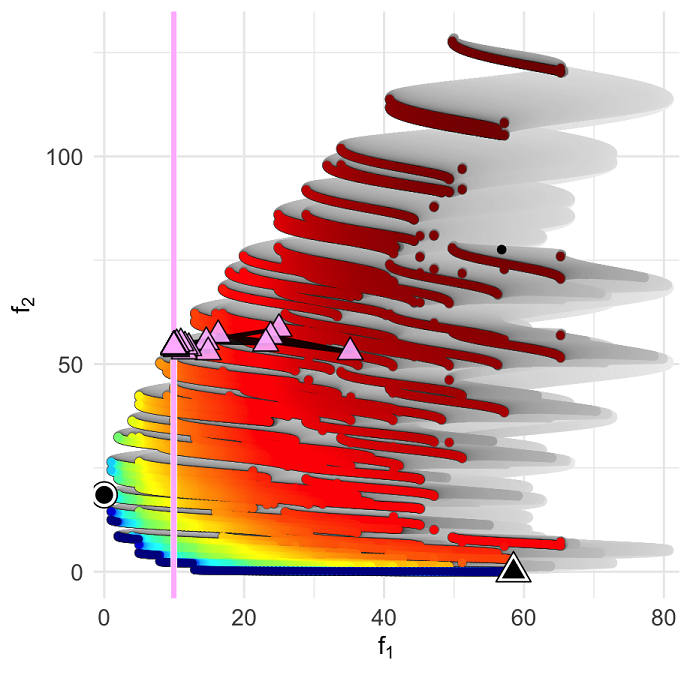}\\[0.25em]
    \includegraphics[width=0.325\textwidth]{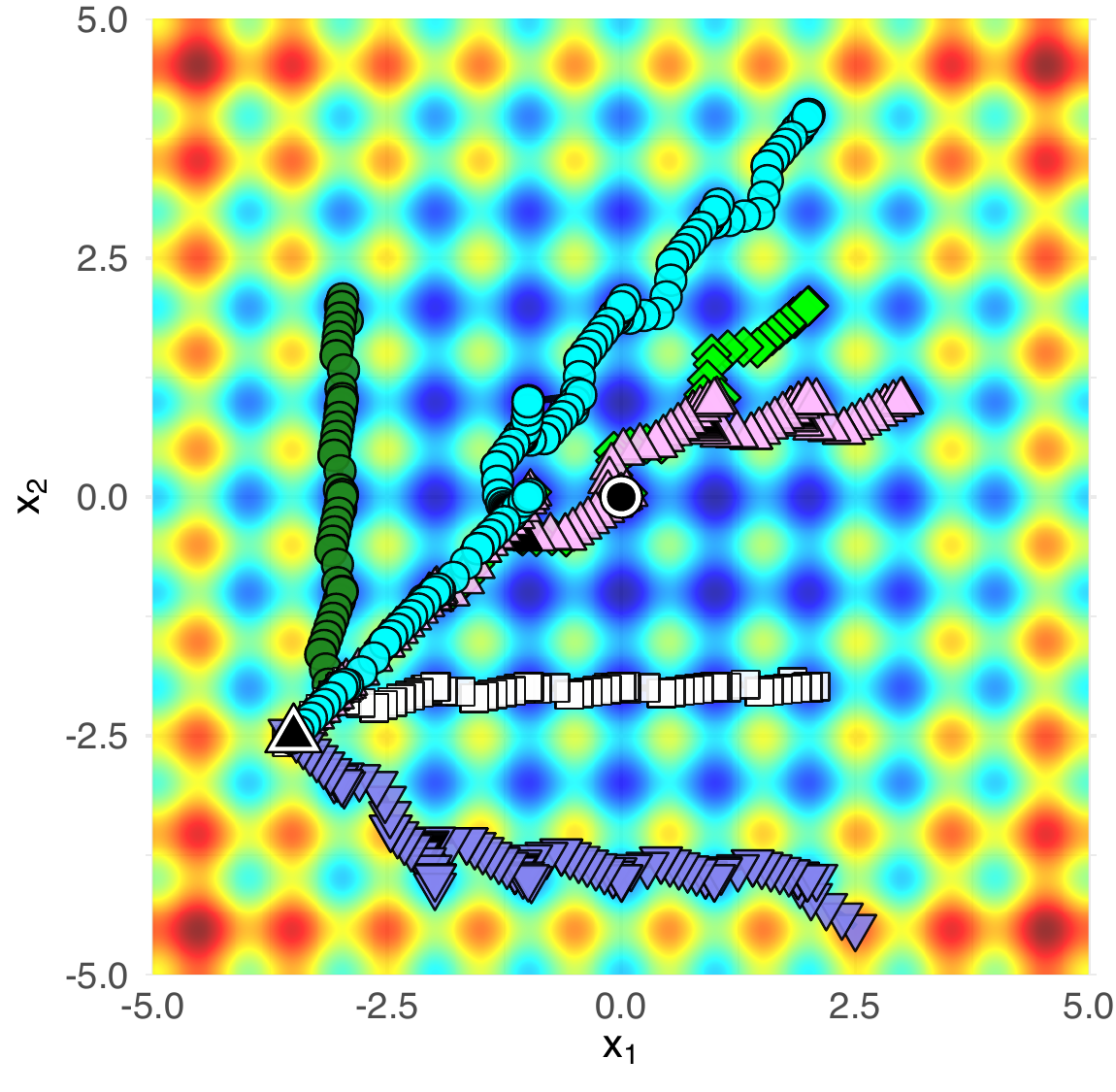}\hfill
    \includegraphics[width=0.325\textwidth]{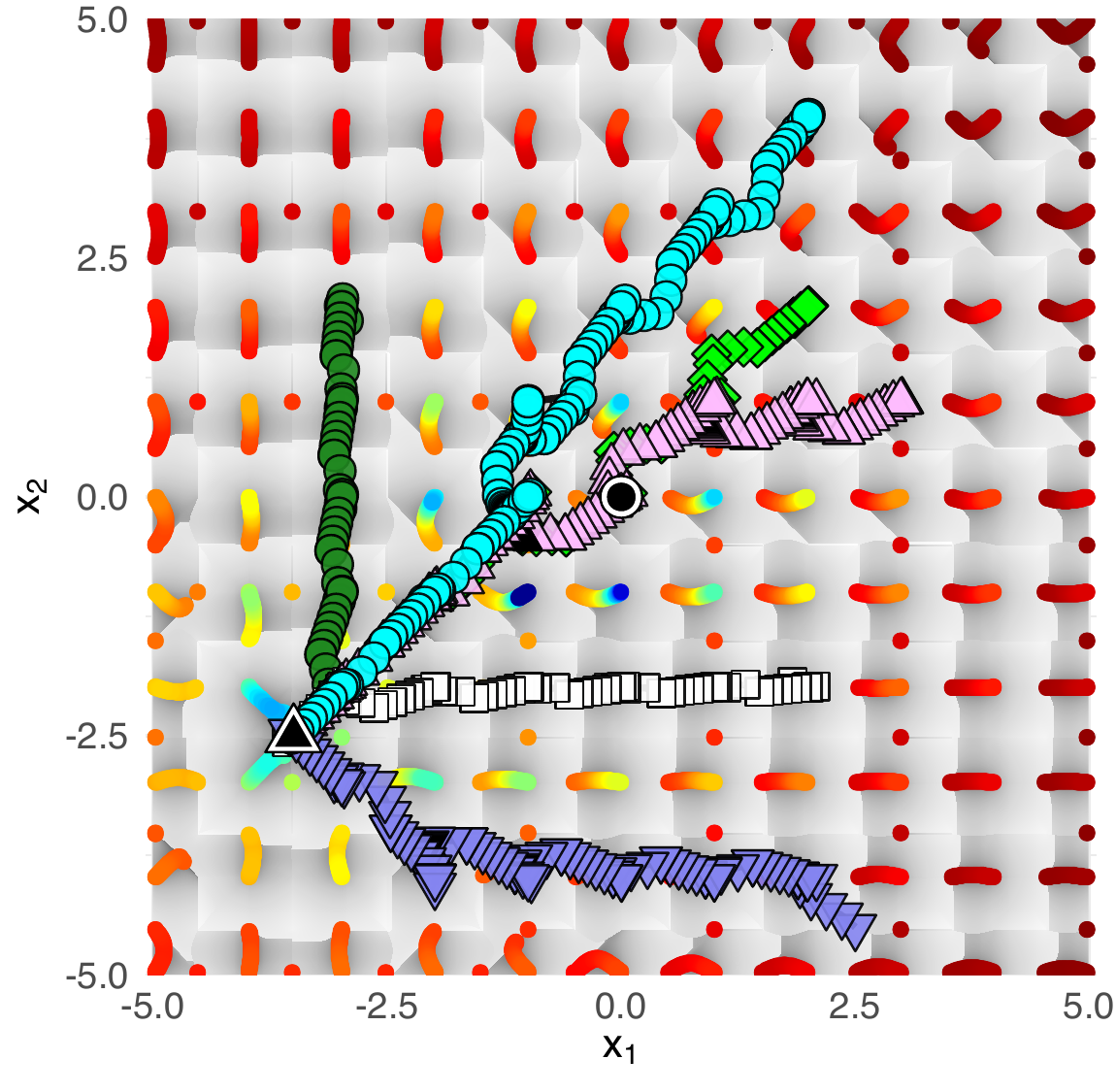}\hfill
    \includegraphics[width=0.325\textwidth]{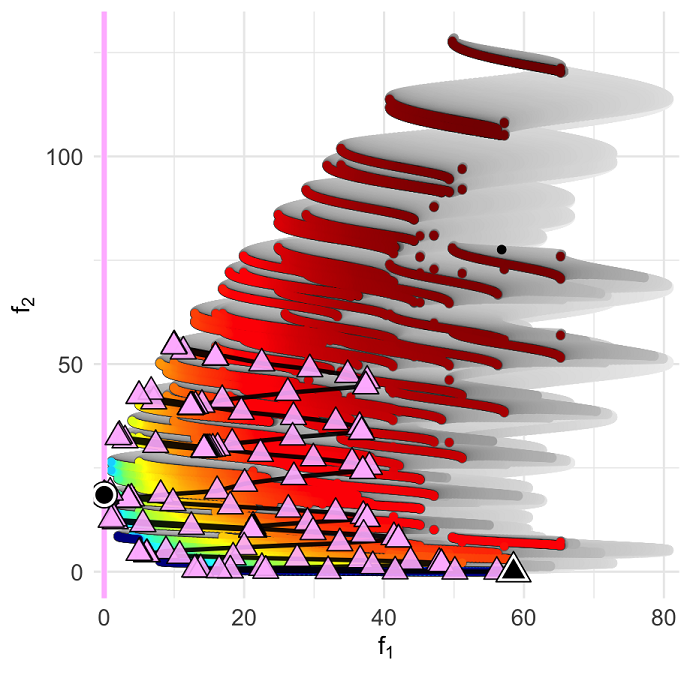}
    \caption{Six runs of Nelder-Mead (top) and SO-MOGSA (bottom) on the Rastrigin function ($f_1$). Search behavior is shown in the single-objective (left) and multi-objectivized search spaces (center).
    Each run has its own starting position identified by a color-shape-combination (\protect\runA, \protect\runB, \protect\runC, \protect\runD, \protect\runE, \protect\runF) Global optima of $f_1$ (\protect\optA) and the sphere (\protect\optB) are located in $(0,0)$ and $(-3.5, -2.5)$, respectively. On the right, we show the objective space for \protect\runD\ and, as vertical line, the final quality w.r.t $f_1$.}
    \label{fig:rastrigin}
\end{figure*}

For visualization, we employ projections of the landscape for the single-objective problem $f_1$ into decision space as well as PLOT landscapes\cite{SchaepermeierGK2020} of the same problems (therein comprising $f_2$ as second objective) to complement the observations with the multiobjectivized view. The search path of both Nelder-Mead and SO-MOGSA is provided as overlays resulting in optimization pathways that augment the respective (single-objective or multiobjectivized) view. In addition, we provide a visualization of the multi-objective objective space of the transformed problem comprising all local efficient fronts and the search path. Figure~\ref{fig:rastrigin} shows the described views for the highly multimodal Rastrigin problem~\cite{HoffBaeck91}. While the left figure is the classical single-objective perspective on the decision space, the middle and right-hand sub-figures depict the multiobjectivized perspective which is exploited by SO-MOGSA. 
As expected, the Nelder-Mead approach (top row) is not able to leave the local optimum, which is nearest to its starting point. The vertical line in the plot of the objective space highlights the best quality of the reached solution when we consider the pink starting point w.r.t. $f_1$. Clearly, as Nelder-Mead stagnates in a local optimum of $f_1$ it also stagnates at a dominated local front in the MO perspective.

In contrast, SO-MOGSA is able to leave the single objective local optima as described in our concept by following the multi-objective gradient descent path and the locally efficient sets as direct connections to neighboring basins of attraction. The depicted paths in Figure~\ref{fig:rastrigin} show that SO-MOGSA descents to the optimum of $f_2$ and on its way passes from one local optimum (in the single-objective perspective) to another. For two out of six starting points in our case study these paths lead through the global optimum. Such a case is also depicted by the sub-figure of the MO objective space: we can observe how the search path descents along local fronts. Thereby, it passes better solutions for $f_1$ and pushes the vertical line towards the global optimal value -- virtually closing a gap between the best solution (at $f_1(x) = 0$ for Rastrigin) and the best yet visited solution.

As for all local search mechanisms (and evident from Fig.~\ref{fig:rastrigin}), the solution quality delivered by SO-MOGSA depends on the starting point. However, we expect SO-MOGSA to reach better regions of the decision space in average due to the capability of escaping local optima. To express this property in a quantitative way, we compare the normalized quality (performance) gap that is closed by the best local search result $x_b$ w.r.t. $f_1$. Specifically, we compute 
\[ g_{LS}(x_s) = \frac{|f_1(x_b) - f_1(x_s)|}{|f_1(x^*) - f_1(x_s)|} \]
for each considered starting point $x_s$ where $LS$ is the specific local search and $x^*$ is the known global optimum.

In Table~\ref{tbl:results} we show the results for this performance measure for all six starting points and regarding Nelder-Mead and SO-MOGSA, respectively. For the Rastrigin function, the values resemble the observations in Figure~\ref{fig:rastrigin}. While over all runs, SO-MOGSA has an average quality gain of about 75\%, Nelder-Mead realizes only an average gain of about 10\%. Starting in R5, however, also demonstrates the weakness of this measure. For R5, Nelder-Mead realizes a moderate gain, while in Figure~\ref{fig:rastrigin}, Nelder-Mead seems to get stuck. The gain is merely realized because R5 is the only starting point that is located on a local maximum. This leads to a descent for SO-MOGSA and Nelder-Mead alike and realizes a baseline gain for both approaches. Afterwards, only SO-MOGSA is able to escape the local optimum and to close the quality gap further.

\setlength{\tabcolsep}{1.7pt}
\begin{table}[t]
\caption{Ratio of performance gap (between starting point and optimum of $f_1$) as closed by the respective algorithm.}
\label{tbl:results}
\centering
\begin{tabular}{lrrrrrrr}
  \toprule
 & \!R1 \protect\runA\! & \!R2 \protect\runB\! & \!R3 \protect\runC\! & \!R4 \protect\runD\! & \!R5 \protect\runE\! & \!R6 \protect\runF\! & $\varnothing$ \\ 
  \midrule
  \multicolumn{8}{l}{\emph{Rastrigin $\rightarrow$ see Fig.~\ref{fig:rastrigin}}}\\
  Nelder-Mead & 0.5\% & 0.5\% & 0.5\% & 0.5\% & 62.6\% & 0.5\% & 10.9\% \\ 
  SO-MOGSA & \bf 31.1\% & \bf 50.3\% & \bf 100.0\% & \bf 100.0\% & \bf 76.1\% & \bf 95.0\% & \bf 75.4\% \\ 
  \midrule
  \multicolumn{8}{l}{\emph{Gallagher's 21 Peaks (Instance 3) $\rightarrow$ see Fig.~\ref{fig:heatmaps_gallagher_vs2}}}\\
  Nelder-Mead & 79.7\% & 88.7\% & 98.8\% & 79.1\% & \bf 100.0\% & 99.0\% & 90.9\% \\ 
  SO-MOGSA & 79.7\% & 88.7\% & 98.8\% & \bf 98.3\% & 99.9\% & \bf 100.0\% & \bf 94.3\% \\ 
  \midrule
  \multicolumn{8}{l}{\emph{Gallagher's 101 Peaks (Instance 1) $\rightarrow$ see Fig.~\ref{fig:heatmaps_gallagher_vs3}}}\\
  Nelder-Mead & 59.2\% & 80.2\% & \bf 100.0\% & 2.5\% & \bf85.8\% & 68.2\% & 66.0\% \\ 
  SO-MOGSA & 59.2\% & \bf 100.0\% & \bf 100.0\% & \bf 81.8\% & 84.5\% & \bf 100.0\% & \bf 87.6\% \\ 
   \bottomrule
\end{tabular}
\end{table}

\begin{figure*}[t]
    \centering
    \includegraphics[width=0.325\textwidth]{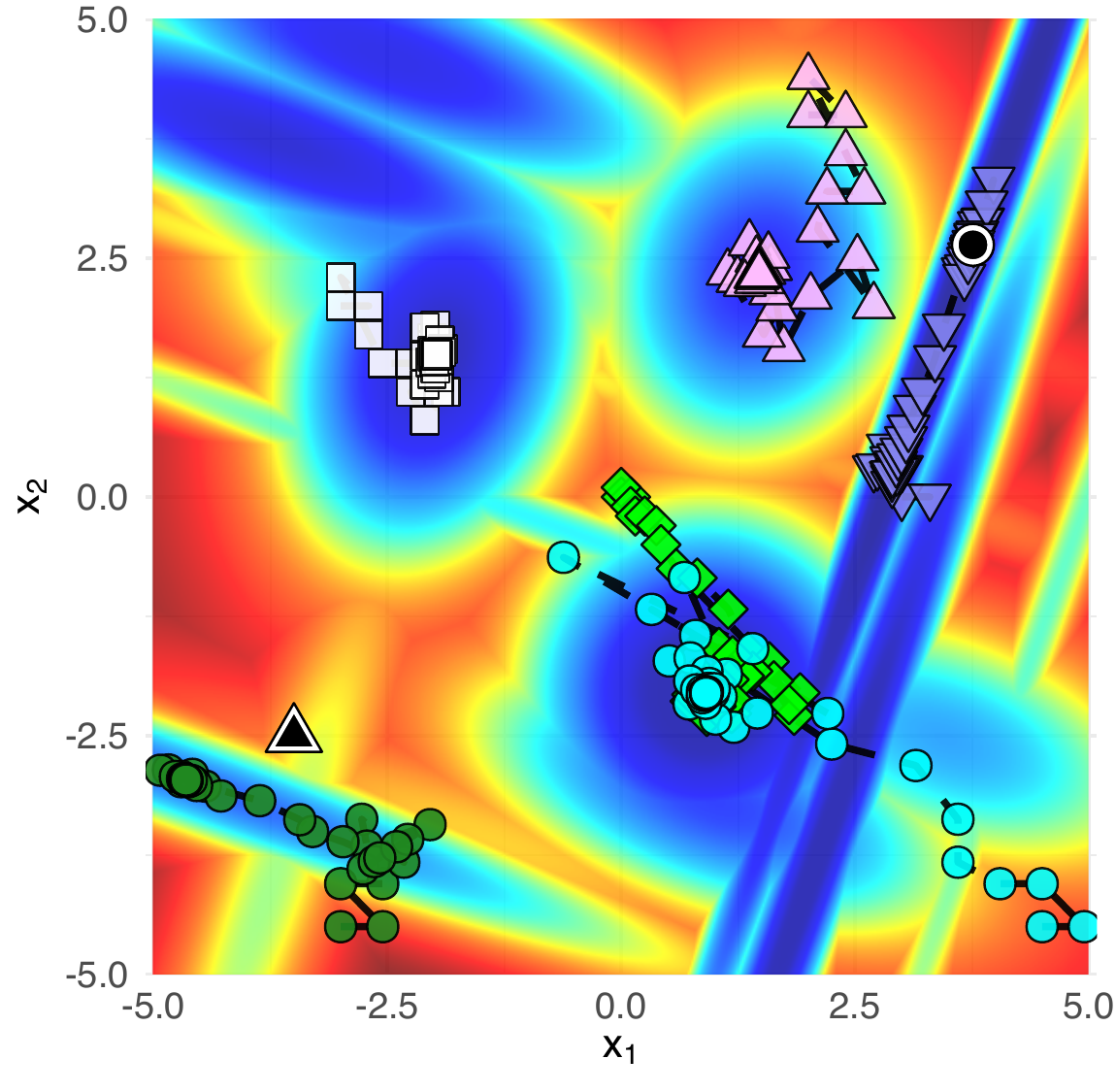}\hfill
    \includegraphics[width=0.325\textwidth]{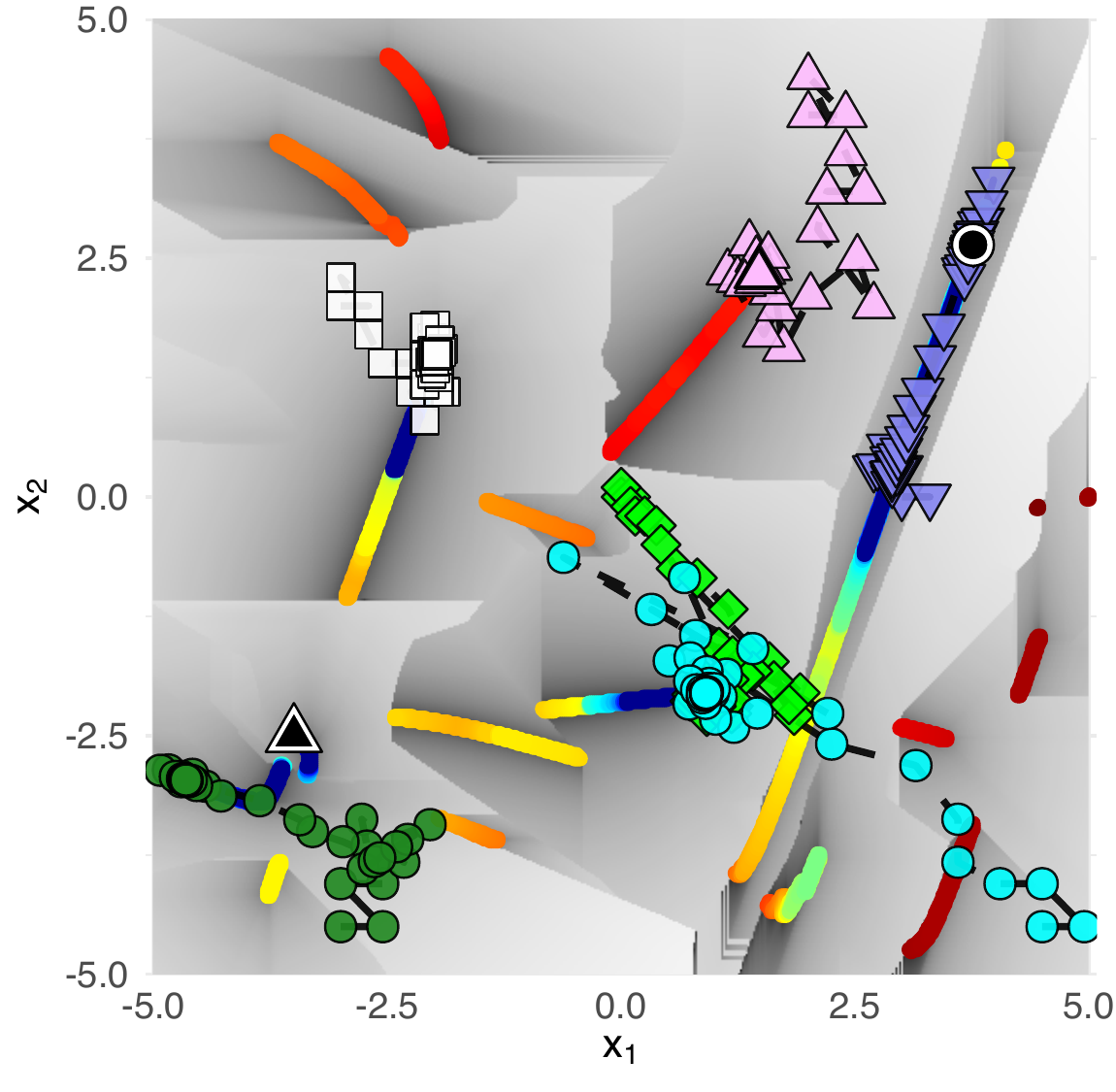}\hfill
    \includegraphics[width=0.325\textwidth]{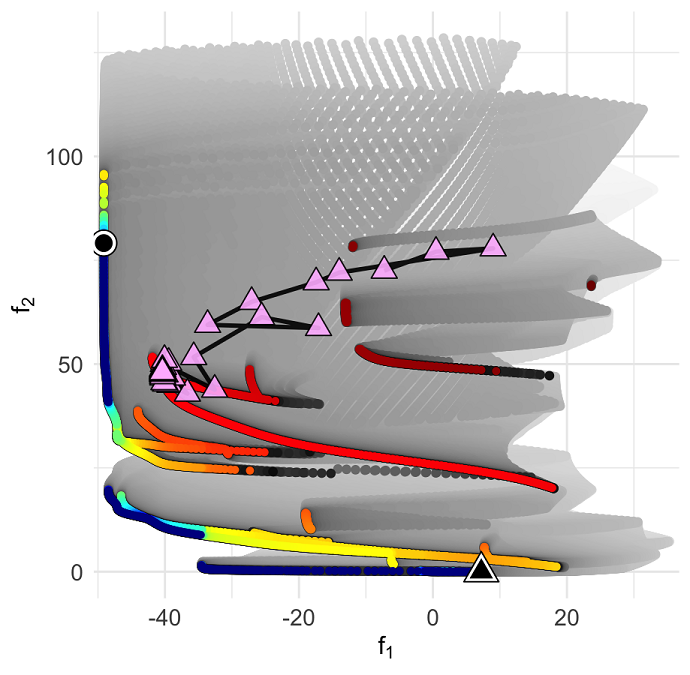}\\[0.25em]
    \includegraphics[width=0.325\textwidth]{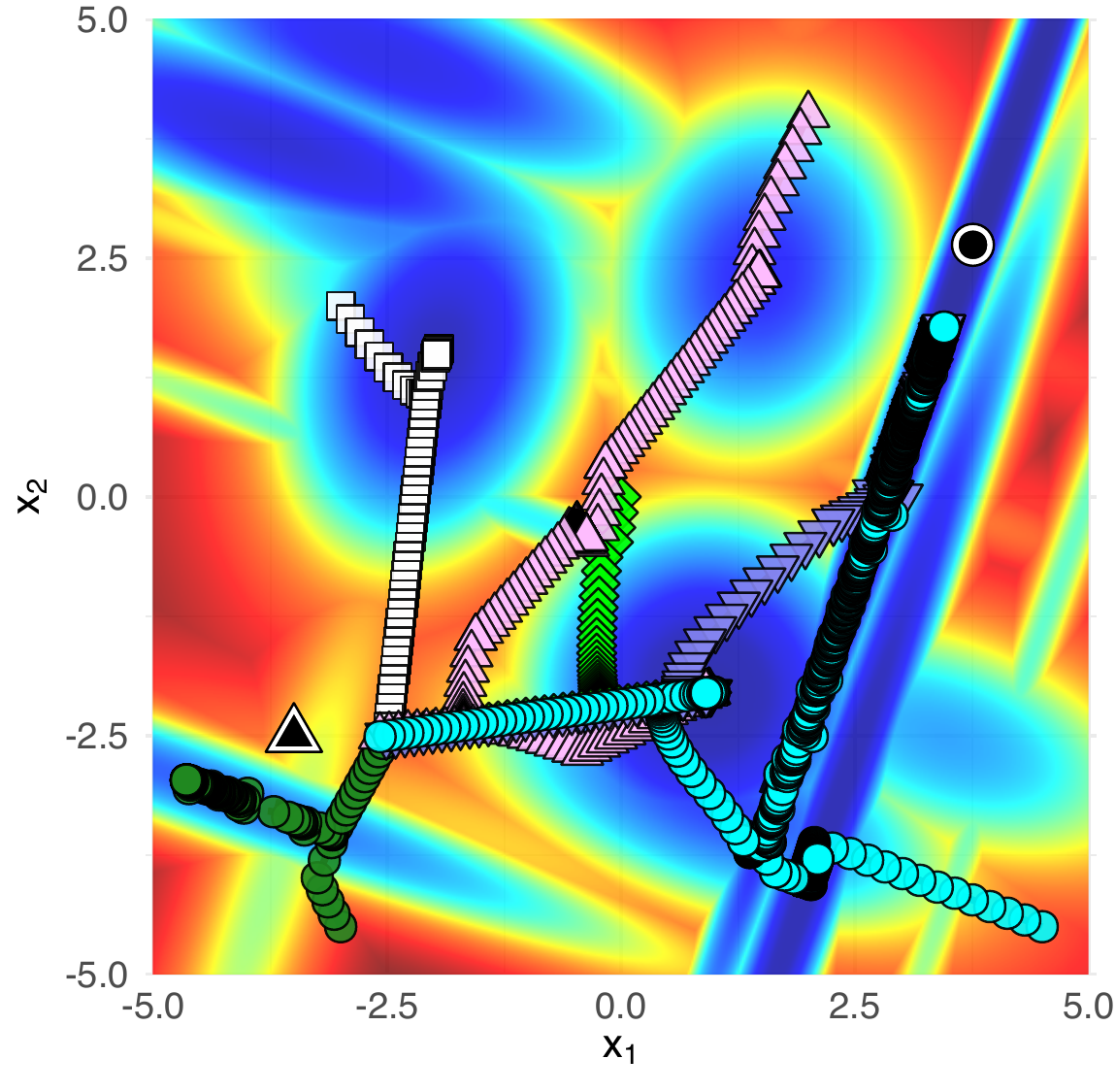}\hfill
    \includegraphics[width=0.325\textwidth]{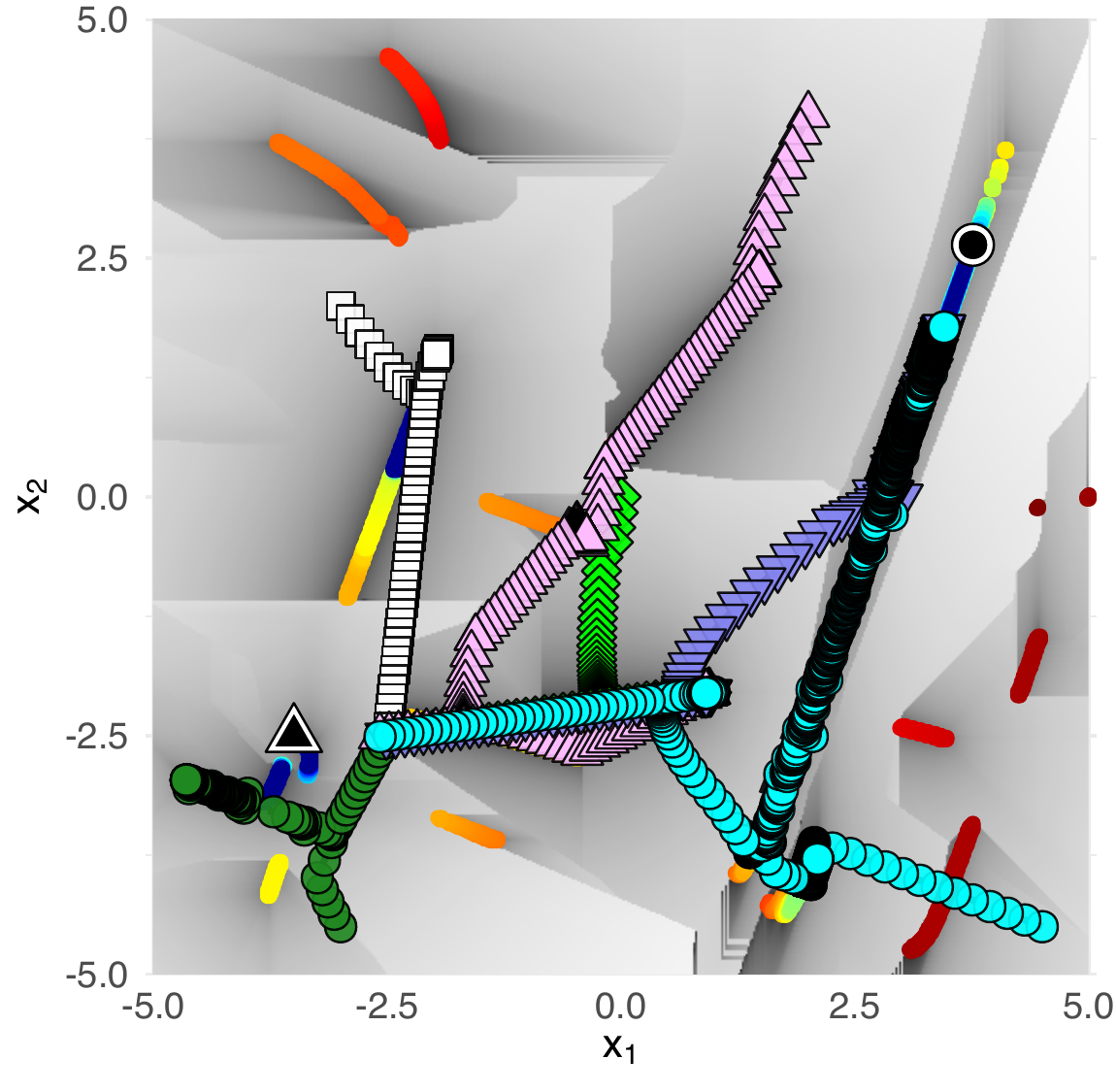}\hfill
    \includegraphics[width=0.325\textwidth]{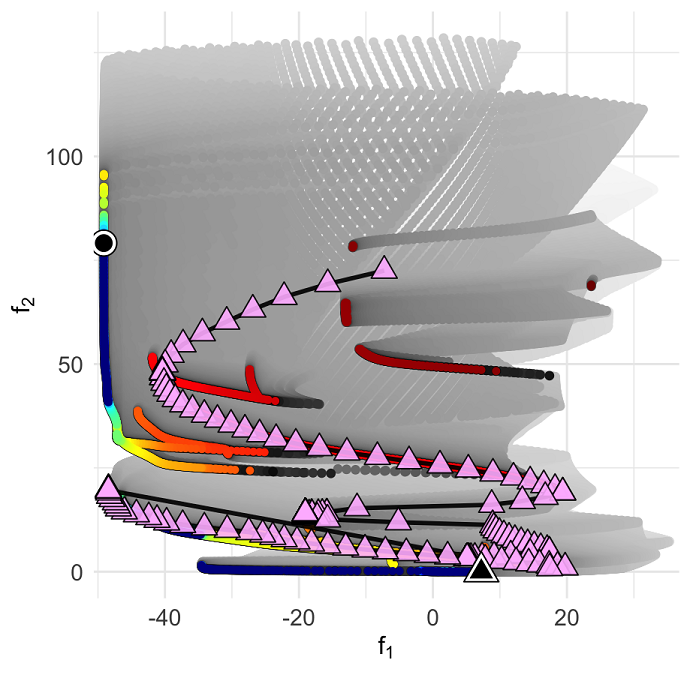}
    \caption{Six exemplary runs of Nelder-Mead (top) and SO-MOGSA (bottom) on an instance of Gallagher's 21 peaks (BBOB function 22)~\cite{hansen2009}. The optimum of the sphere (\protect\optB) is placed in $(-3.5, -2.5)$.}
    \label{fig:heatmaps_gallagher_vs2}
\end{figure*}

\begin{figure*}[t]
    \includegraphics[width=0.325\textwidth]{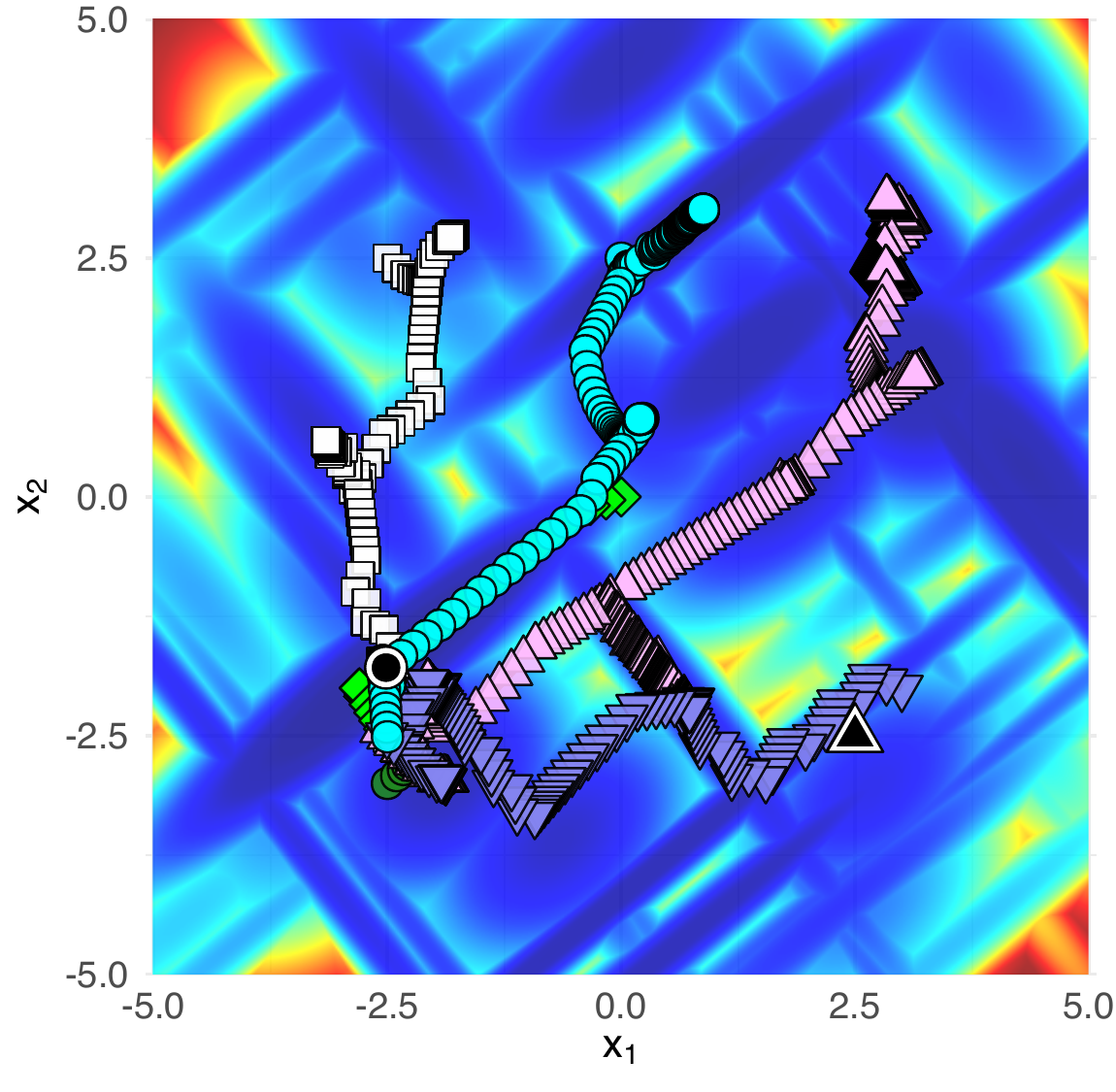}\hfill
    \includegraphics[width=0.325\textwidth]{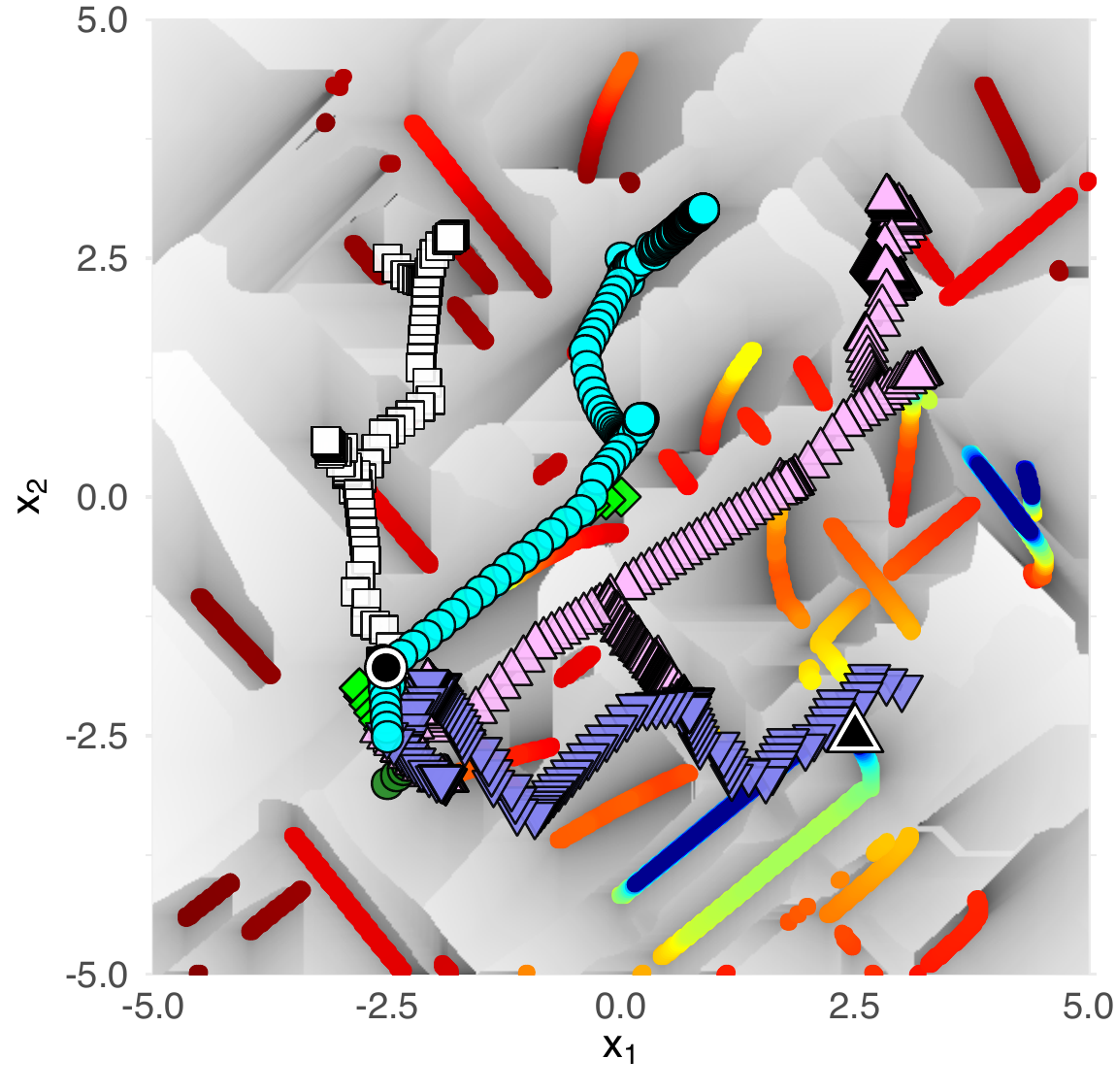}\hfill
    \includegraphics[width=0.325\textwidth]{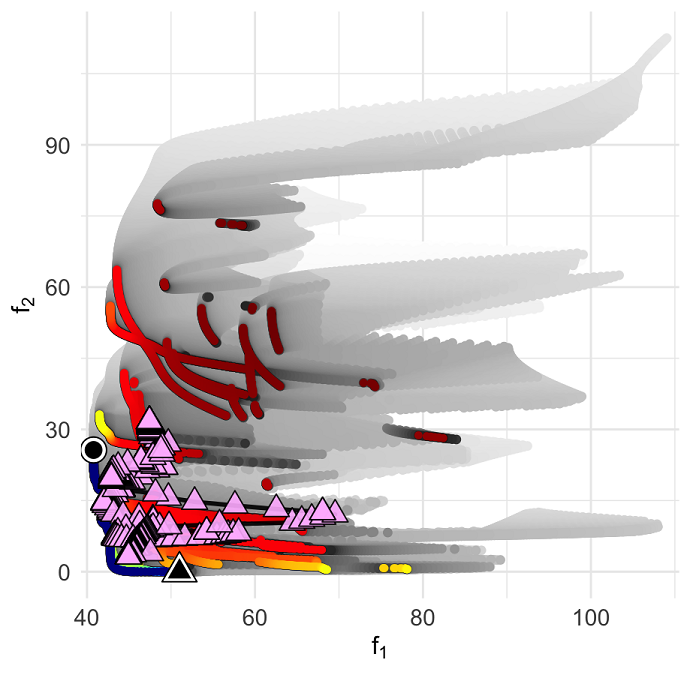}
    \caption{Illustration of six exemplary runs of SO-MOGSA on an instance of Gallagher's 101 peaks (BBOB function 21)~\cite{hansen2009}. The optimum of the sphere (\protect\optB) is placed in $(2.5, -2.5)$.}
    \label{fig:heatmaps_gallagher_vs3}
\end{figure*}

To confirm the working principle for further complex multimodal problems, we also include Gallagher's 21 and 101 peaks problems~\cite{hansen2009} into this study, see Figures~\ref{fig:heatmaps_gallagher_vs2} and~\ref{fig:heatmaps_gallagher_vs3}. 
The figures as well as the observed individual and average quality gain in Table~\ref{tbl:results} confirm that local optima are no traps for SO-MOGSA, in principle. Although SO-MOGSA is a purely gradient-based strategy, it is often able to close the performance gap between starting point and the global optimal value better than classical local search. However, even more important for the general understanding of the benefit of multiobjectivization, these experiments provide evidence that additional objectives and the integration of MO landscape characteristics can help local search in escaping local optima. The visualization applied here proves that local efficient sets can be pathways for reaching neighboring local basins, which are traps for classical local search in the single-objective domain.



\section{Conclusion}\label{sec:concl}
This work contributed in two ways to the field. On the one hand, we provided a new concept of gradient-based local search, which exploits characteristics of multi-objective landscapes and helps local search to escape local optima via locally efficient sets. Second, we delivered a visually accessible explanation how and why the proposed method benefits from multiobjectivization - something often claimed but not sufficiently explained. We believe that this is only the onset of further research addressing limitations of the current approach and extending insights into multiobjectivized problems' behaviour. Further research may comprise a more efficient implementation of SO-MOGSA, rigorous performance evaluation, or effective integration into meta-heuristics like evolutionary algorithms. Additionally, specific questions on the parameterization and configuration of the helper function $f_2$ during multiobjectivization -- e.g. where to locate the local optimum -- are of great importance.


\subsection*{Acknowledgments}
The authors acknowledge support by the \href{https://www.ercis.org}{\textit{European Research Center for Information Systems (ERCIS)} and the LIACS, Leiden, NL}.


\bibliography{arxiv}
\bibliographystyle{abbrv}
\end{document}